\documentclass[letterpaper]{article} 
\usepackage[preprint]{aaai2027}  
\usepackage[hyphens]{url}  
\usepackage{graphicx} 
\usepackage{natbib}  
\usepackage{caption} 
\usepackage{makecell}
\usepackage{algorithm}
\usepackage{algorithmic}
\usepackage{newfloat}
\usepackage{listings}
\DeclareCaptionStyle{ruled}{labelfont=normalfont,labelsep=colon,strut=off} 
\floatstyle{ruled}
\newfloat{listing}{tb}{lst}{}
\floatname{listing}{Listing}

\usepackage{booktabs}
\usepackage{colortbl}
\usepackage{multirow}

\usepackage{amsmath}
\usepackage{amssymb}
\usepackage{amsthm}

\usepackage[capitalise]{cleveref}

\crefname{algorithm}{Algorithm}{Algorithms}
\Crefname{algorithm}{Algorithm}{Algorithms}
\crefname{listing}{Listing}{Listings}
\Crefname{listing}{Listing}{Listings}

\usepackage{tabularx}
\usepackage{array}
\newcolumntype{Y}{>{\raggedright\arraybackslash}X}

\newcommand{\ptdiff}[1]{\textsuperscript{\normalfont(#1)}}

\title{Learning Where Outcomes Change: \\Credit-Addressable Reasoning for Multimodal Geometry}
\author {
    Jiani Guo\textsuperscript{\rm 1,}\equalcontrib,
    Junjie Wang\textsuperscript{\rm 1}\equalcontrib\corresponding,
    Jie Wu\textsuperscript{\rm 1},
    Pengxiang Zhao\textsuperscript{\rm 3},
    Dongdong Zhang\textsuperscript{\rm 2}\corresponding,\\
    Shaohan Huang\textsuperscript{\rm 2},
    Yujiu Yang\textsuperscript{\rm 1}\corresponding,
    Furu Wei\textsuperscript{\rm 2}
}
\affiliations {
    \textsuperscript{\rm 1}Tsinghua University
    \qquad
    \textsuperscript{\rm 2}Microsoft Research
    \qquad
    \textsuperscript{\rm 3}Zhejiang University\\
    guojn26@mails.tsinghua.edu.cn, 
     \qquad
    wangjunjie@sz.tsinghua.edu.cn,\\
     \qquad
    dozhang@microsoft.com,
     \qquad
    yang.yujiu@sz.tsinghua.edu.cn

     \vspace{8pt}
  {\centering \url{https://github.com/gjn12-31/CE-GRPO} \par}
}

\begin{document}

\maketitle

\begin{abstract}
Multimodal geometry reasoning requires VLMs to extract precise visual relations and preserve them through multi-step deduction.
Existing free-form traces obscure the decisions that determine the answer, and trajectory-level reinforcement learning distributes a single terminal signal across the entire response.
We introduce \emph{credit-addressable reasoning}, in which the semantic units exposed during inference also define where learning compares alternatives and assigns credit.
We instantiate this principle with Code-CoT, which retains the diagram, represents visual relations as line-addressable executable code, and organizes reasoning into typed events, and CE-GRPO, which selects event boundaries using structural priors and type-normalized entropy, samples complete continuations from shared prefixes, and converts outcome differences into localized advantages.
Across nine geometry benchmarks, CE-GRPO achieves an average accuracy of $76.04$, outperforming Qwen3-VL-8B and trajectory-level GRPO by $8.09$ and $3.43$ points, respectively.
Its relative advantage increases with the number of intermediate events, demonstrating the value of representation--optimization co-design for long, dependency-heavy multimodal reasoning.
\end{abstract}

\begin{figure}[t]
\centering
\includegraphics[width=0.88\linewidth,height=0.40\textheight]{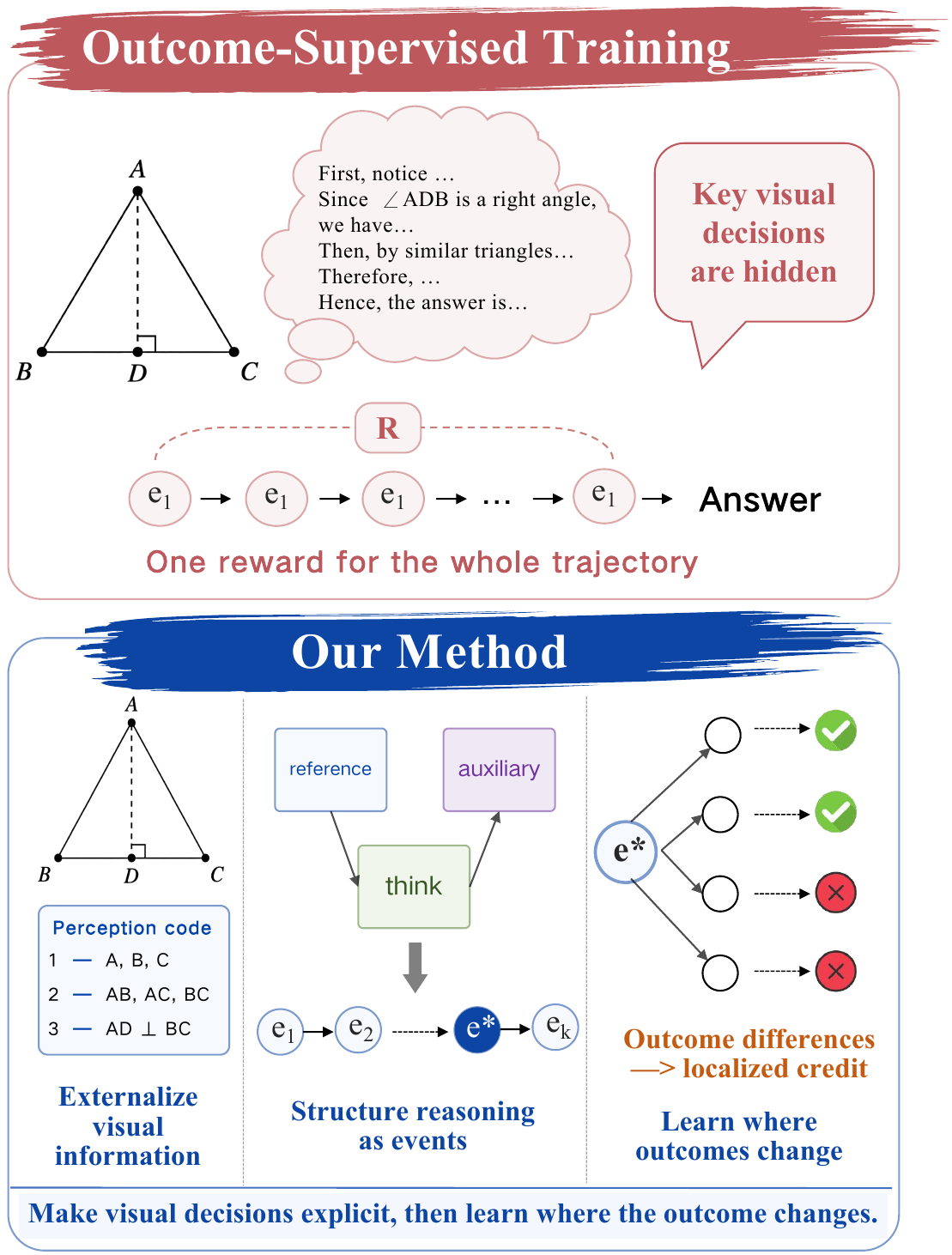}
\caption{Free-form reasoning hides visual decisions and receives one trajectory reward; Code-CoT makes decisions explicit, and CE-GRPO assigns localized credit.}
\label{fig:concept}
\end{figure}

\section{Introduction}
\label{sec:intro}

Recent vision-language models (VLMs) have made substantial progress in mathematical reasoning~\cite{shao2024deepseekmath,bai2025qwen3vl}. 
Geometry, however, remains a stringent test of reliable multimodal reasoning: models must accurately interpret and consistently use diagrammatic relations, while a single error in object binding, angle interpretation, or auxiliary construction can invalidate subsequent deductions~\cite{zhang2024mathverse,kamoi2025visonlyqa,fu2026geolaux}. 
As shown in~\cref{fig:concept}, these critical decisions remain implicit in a free-form reasoning trace, while trajectory-level optimization provides only a single outcome signal for the response as a whole. 
This creates two coupled gaps: a \emph{representation gap}, where decisive choices lack explicit semantic units, and a \emph{credit gap}, where terminal feedback cannot distinguish their individual effects.

Prior work has approached these gaps along two largely separate directions. 
Program-aided methods such as PAL and ToRA organize computation through programs or tools, while formal geometry systems such as AlphaGeometry and GeoTikzBridge translate geometric relations into formal languages or executable representations~\cite{gao2023pal,gou2024tora,trinh2024alphageometry,sun2026geotikzbridge}.
Fine-grained reinforcement-learning methods such as VinePPO, Segment Policy Optimization, GPO, and GRPO-MA instead estimate local advantages at intermediate states, segments, critical steps, or branched thoughts~\cite{kazemnejad2025vineppo,yu2025gpo,wang2026branching}. 
The former expose reasoning structure without making it an optimization unit; the latter refine update locations but still derive semantic decisions from learned values, fixed boundaries, or token statistics. 
What remains missing is a semantic unit shared by reasoning and learning. 
We call this \emph{credit-addressable reasoning}: inference-time decisions directly govern history correction, future comparison, and credit assignment.

\begin{table}[!t]
\centering
\small
\setlength{\tabcolsep}{4pt}
\begin{tabular}{lccccc}
\toprule
Model & $I$ & $C$ & $I{+}C$ & $\Delta$ & $C_{\mathrm{self}}$ \\
\midrule
\multicolumn{6}{l}{\textit{Open models}} \\
InternVL3.5-8B  & 45.5 & 47.0 & \textbf{63.9} & $+18.4$ & 33.4 \\
Qwen2.5-VL-7B   & 49.4 & 51.2 & \textbf{60.6} & $+11.2$ & 34.6 \\
Qwen3.5-9B      & 79.9 & 46.8 & \textbf{85.4} & $+5.5$  & 35.1 \\
\midrule
\multicolumn{6}{l}{\textit{Proprietary models}} \\
Kimi-K2.6       & 88.2 & 87.9 & \textbf{92.6} & $+4.4$ & -- \\
Claude-Opus-4.8 & 87.9 & 86.3 & \textbf{91.1} & $+3.2$ & -- \\
Gemini-3.1-Pro  & 92.3 & 89.1 & \textbf{93.0} & $+0.7$ & -- \\
\bottomrule
\end{tabular}
\caption{
Mean accuracy on the five MathVerse~\cite{zhang2024mathverse} subsets.
$I$: diagram only; $C$: externally generated code only;
$I{+}C$: diagram and external code.
$\Delta=(I{+}C)-I$ is the absolute gain in percentage points.
$C_{\mathrm{self}}$ uses self-generated code only, with the
diagram withheld, and is reported for open models.
}
\label{tab:prestudy}
\end{table}

To identify such a shared semantic unit, we first examine whether code can serve as an internal reasoning space for geometry.
As shown in~\cref{tab:prestudy}, combining the diagram with externally generated code consistently outperforms either modality alone across three open and three proprietary models (details in~\cref{sec:motivation}).
For the three open models, self-generated code trails external code by $11.7$--$16.6$ points under the same code-only setting, identifying reliable code generation as the main bottleneck.
These results suggest two design requirements: code should complement rather than replace the image, and its generation must be explicitly learned.
We therefore introduce \textbf{Code-CoT}, which retains the original image, produces line-addressable executable Matplotlib perception code, and organizes subsequent reasoning into \texttt{reference}, \texttt{auxiliary}, \texttt{coordinate}, and \texttt{think} events.
Each event is both a checkable geometric operation and an addressable unit of credit.

Making decisions addressable does not by itself make their credit localizable.
Standard Group Relative Policy Optimization (GRPO)~\cite{shao2024deepseekmath} compares complete responses from the original problem and assigns the same group-relative advantage throughout each response, treating an event-structured Code-CoT trace as a flat trajectory.
We introduce \textbf{Critical-Event Group Relative Policy Optimization (CE-GRPO)}, where a \emph{critical event} is operationally defined as an event boundary whose alternative continuations under the same prefix yield different terminal outcomes.
CE-GRPO selects candidate events using a structural prior and type-normalized entropy, then fixes the image, question, and complete prefix before each candidate and samples multiple continuations through the final answer, as shown in the lower panel of~\cref{fig:overview}.
Each continuation receives the same terminal reward, while the shared prefix is excluded from the policy loss, so the group-relative advantage updates only the regenerated event and its downstream consequences.
Structure provides semantic branch points, entropy allocates the branching budget, and reward variation determines whether a candidate yields a useful credit signal.
If all continuations receive the same reward, the group contributes no update, making an uninformative selection computationally wasteful rather than harmful.

Across nine geometry benchmarks, CE-GRPO achieves an average accuracy of $76.04$, outperforming the native Qwen3-VL-8B~\cite{bai2025qwen3vl} backbone, Code-CoT SFT, and trajectory-level GRPO by $8.09$, $6.49$, and $3.43$ points, respectively, and improving over the backbone on all nine benchmarks.
On validly terminated responses, it retains a $3.91$-point average gain over trajectory-level GRPO, confirming improvements in solution quality rather than protocol compliance.
Offline analysis further shows that structural selection identifies outcome-changing events about $30\%$ more often than random selection, while the CE-GRPO margin grows by $3.77$ points per additional intermediate event ($r=0.866$, exact $p=0.0016$).
Together, these results validate credit-addressable reasoning, particularly for solutions involving longer chains of intermediate decisions.

Our contributions are threefold:
\begin{itemize}
    \item We formulate \emph{credit-addressable reasoning}, where the semantic units exposed during inference also define where optimization compares alternatives and assigns credit; a controlled six-model study motivates code as such a shared representation.
    \item We introduce \textbf{Code-CoT}, which represents diagram perception and geometric operations as executable, addressable events, and \textbf{CE-GRPO}, which compares complete futures from shared prefixes to localize terminal credit without process annotations or an auxiliary value model.

    \item Across nine geometry benchmarks, our method improves over the backbone on every benchmark and outperforms trajectory-level GRPO on average; closed-only evaluation, selector validation, and event-count analysis further support the mechanism of event-localized credit.
\end{itemize}

\section{Related Work}

\noindent\textbf{Executable representations for multimodal geometry reasoning.}
Program-aided methods express reasoning as executable programs or tool calls~\cite{gao2023pal,chen2023programofthoughts,gou2024tora}, while geometry systems convert diagrams into formal relations or code~\cite{lu2021intergps,trinh2024alphageometry,sun2026geotikzbridge,wang2026geoparsing}.
Recent work further uses rendered code, visual actions, and dynamic constructions~\cite{wang2025visuothink,duan2026codeplotcot,shi2026mathcanvas,su2025pixelreasoner,wei2026geointr1}.
Unlike these external or intermediate representations, Code-CoT integrates the original diagram and line-addressable code into typed reasoning events, enabling both verification and policy optimization.

\noindent\textbf{Fine-grained credit assignment for reasoning.}
PPO, GRPO, and DAPO optimize complete responses with sequence-level or terminal rewards~\cite{schulman2017ppo,shao2024deepseekmath,yu2025dapo}.
Fine-grained methods assign credit to intermediate steps or high-entropy tokens~\cite{kazemnejad2025vineppo,guo2025spo,yu2025gpo,samanta2026creditassignmentresetslanguage,wang2026branching,wang2025highentropy}, while multimodal variants use visual perturbations or auxiliary rewards~\cite{wang2026papo,wang2026vgpo,yu2026cfpo,wang2026geometryzero,guo2025geovlmath}.
CE-GRPO instead branches at Code-CoT event boundaries, allocates computation by type-normalized entropy, and derives local credit from terminal outcome differences without process labels, value models, or task-specific rewards.

\section{Motivation: Code as a Reasoning Space}
\label{sec:motivation}

\begin{figure}[!t]
\centering
\includegraphics[width=\linewidth]{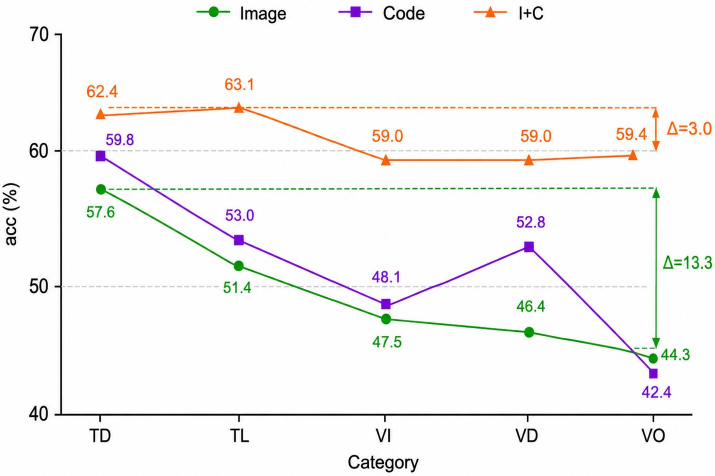}
\caption{
Qwen2.5-VL-7B accuracy across the five MathVerse subsets under diagram-only ($I$), external-code-only ($C$), and combined diagram--code ($I{+}C$) inputs.
Combining both modalities reduces the Text-Dominant--Vision-Only gap from $13.3$ to $3.0$ points.
}
\label{fig:modality-gap}
\end{figure}

\begin{figure*}[!t]
\centering
\includegraphics[width=0.98\textwidth]{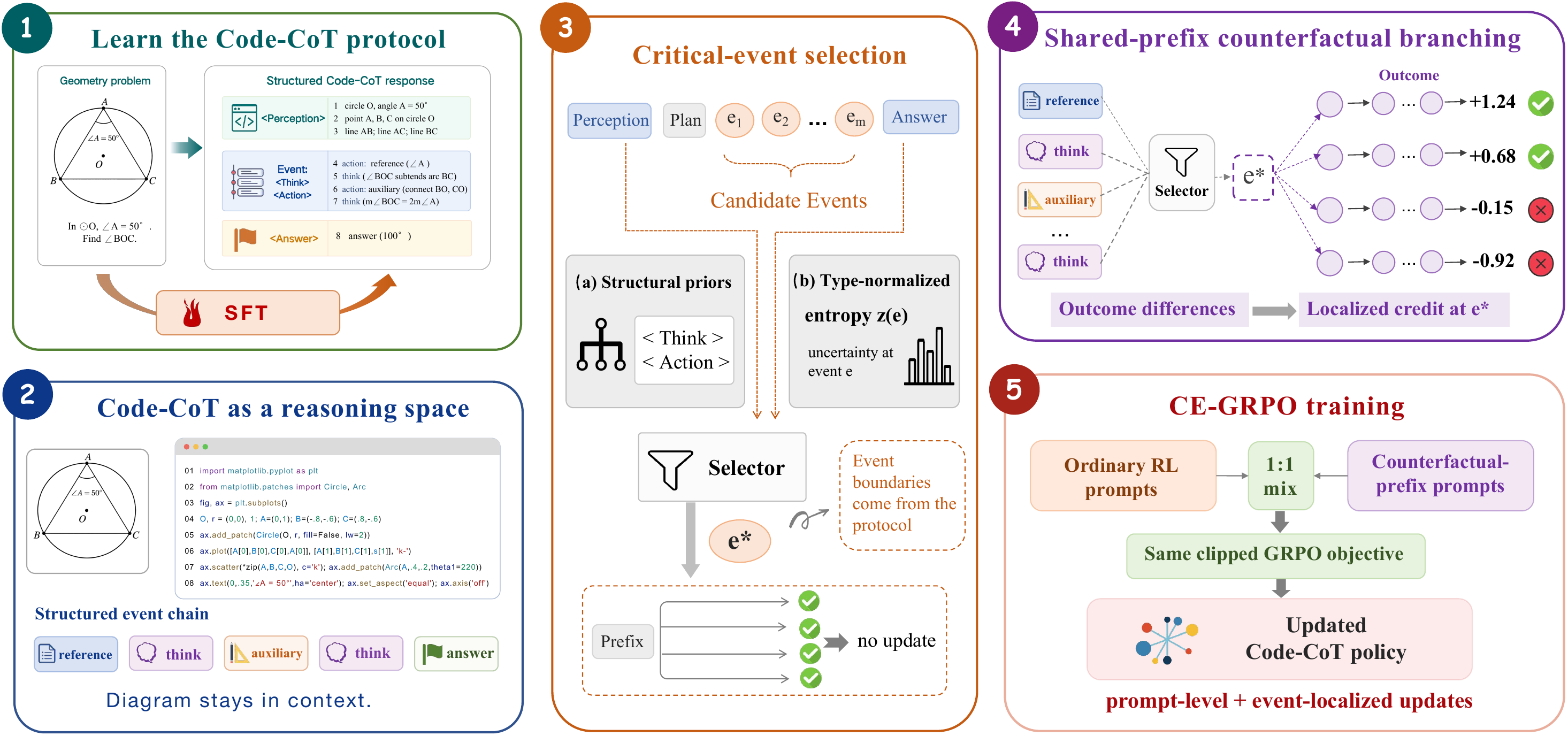}
\caption{
Overview of Code-CoT with CE-GRPO.
(1) Supervised fine-tuning installs the Code-CoT protocol from curated traces.
(2) The model retains the diagram and reasons through executable perception code and typed events.
(3) Structural priors and type-normalized entropy select candidate events.
(4) Shared-prefix branching compares complete continuations and converts outcome differences into localized credit.
(5) Ordinary and counterfactual-prefix prompts are mixed under the same objective.
}
\label{fig:overview}
\end{figure*}

\noindent\textbf{Controlled study.}
To determine whether code can serve as the shared semantic unit sought above, we study how its utility depends on the available modality and the source of the code.
MathVerse~\citep{zhang2024mathverse} presents each problem in five variants that progressively shift information from text to the diagram.
We compare diagram only ($I$), code generated externally by Gemini-3.1-Pro-Preview only ($C$), both ($I{+}C$), and self-generated code only ($C_{\mathrm{self}}$).
The diagram is withheld under both code-only conditions, so $C$ and $C_{\mathrm{self}}$ differ only in code source.
We evaluate $I$, $C$, and $I{+}C$ on three open and three proprietary models, and additionally evaluate $C_{\mathrm{self}}$ on the open models.
\Cref{tab:prestudy} reports the cross-model means, while \cref{fig:modality-gap} shows the effect of increasing visual dependence.

\noindent\textbf{Findings.}
Three patterns emerge.
First, code complements rather than replaces the diagram: $I{+}C$ performs best for all six models and reduces the Qwen2.5-VL-7B Text-Dominant--Vision-Only gap from $13.3$ to $3.0$ points.
Second, code is primarily compensatory: the gain $\Delta=(I{+}C)-I$ generally decreases as diagram-only accuracy increases, indicating greater utility when native visual reasoning is weaker~\citep{namgoong2026auxiliary}.
Third, reliable generation is the bottleneck: under the same code-only setting, $C_{\mathrm{self}}$ trails $C$ by $11.7$--$16.6$ points for every open model.
This comparison isolates the effect of code source, although it does not separate perception errors from errors in translating a correct perception into code.

\noindent\textbf{Design implications.}
These findings impose three requirements: code should remain alongside the diagram, be produced by the reasoning model itself, and be explicitly learned rather than merely prompted.
We therefore treat code as a persistent, line-addressable reasoning space rather than an external tool or fixed intermediate artifact.
Code-CoT realizes this design by expressing visual retrieval, geometric construction, coordinate setup, and deduction as explicit events, providing the semantic units later shared by programmatic verification and localized policy optimization.

\section{Method}
\label{sec:method}

\subsection{Overview}
\label{sec:method:overview}

Given an image--question pair $x=(I,Q)$, our policy generates a complete Code-CoT response
\begin{equation}
Y=\left\langle C,P,(e_\ell)_{\ell=1}^{L},F\right\rangle,
\qquad e_\ell\in\mathcal{E},
\label{eq:codecot_output}
\end{equation}
where $C$ is line-addressable executable perception code, $P$ is a solution plan, $F$ is the final answer, and $\mathcal{E}=\{\texttt{think},\texttt{reference},\texttt{auxiliary},\texttt{coordinate}\}$ defines the intermediate event types.

As shown in~\cref{fig:overview}, the framework proceeds in five stages.
First, supervised fine-tuning installs the Code-CoT protocol from curated traces.
The resulting policy retains the diagram and reasons through executable perception code and typed events.
CE-GRPO then selects candidate events using structural priors and type-normalized entropy, branches multiple complete futures from a shared prefix, and converts their outcome differences into event-conditioned credit.
Finally, ordinary and counterfactual-prefix prompts are jointly optimized under the same GRPO objective.
The event boundaries exposed during reasoning therefore also define the units used during optimization.

At inference, the model autoregressively generates the perception code, reasoning events, and final answer in a single response, without an external solver or test-time branching.

\subsection{Code-CoT: A Credit-Addressable Reasoning Space}
\label{sec:method:codecot}

Steps~1--2 in~\cref{fig:overview} establish the Code-CoT policy:
Step~1 installs the protocol through curated supervision, and Step~2 realizes it as a persistent reasoning space.

\noindent\textbf{Persistent executable representation.}
The model converts the input diagram into line-numbered, executable Matplotlib perception code $C$, while retaining the original image throughout the response.
The code therefore complements rather than replaces the visual input, providing a line-addressable interface for retrieving diagram facts and extending the working geometry.
The same policy generates the code, reasoning trace, and final answer within a single response, without an external geometry solver.

\noindent\textbf{Typed event reasoning.}
After producing $C$ and the solution plan $P$, the model interleaves \texttt{think} events with three typed actions before returning final answer $F$.
\texttt{reference} retrieves code lines supporting a diagram fact, \texttt{auxiliary} extends the working geometry, and \texttt{coordinate} establishes an executable coordinate frame.
Subsequent \texttt{think} events reason over the state, while actions are introduced only when required by the solution.

\noindent\textbf{Verification and credit addressability.}

Programmatic checkers validate response structure, code executability, line-grounded references, and typed actions; see Appendix A.1.
Protocol tags deterministically partition the trace into complete \texttt{think} and action events.

Each event is therefore \emph{addressable}: CE-GRPO can recover its pre-event prefix and branch alternative continuations from the same state.
Therefore, Code-CoT jointly supports programmatic verification and event-conditioned policy optimization.

\subsection{Learning Code-Grounded Reasoning}
\label{sec:method:learning}

\noindent\textbf{Supervised initialization.}
Step~1 in~\cref{fig:overview} internalizes Code-CoT through supervised fine-tuning on $18{,}302$ quality-controlled traces.
Given an image--question pair $x=(I,Q)$, the policy learns the complete mapping in~\cref{eq:codecot_output}, jointly generating perception code, a solution plan, typed events, and the final answer.
Each retained trace passes structural, execution, grounding, action-validity, and answer-correctness checks.
Reinforcement learning then uses a separate deduplicated problem pool, on which the current policy generates the full Code-CoT response online.
See Appendix A.2 for data construction and filtering.

\noindent\textbf{Programmatic reward.}
We evaluate a generated response $y$ using
\begin{equation}
R(x,y)=
\left\{
\begin{aligned}
-1,
& \quad y\notin\mathcal{V},\\
\operatorname{clip}\!\left(
c(y)+\lambda a(y)-\Omega(y),-1,1.3
\right),
& \quad y\in\mathcal{V},
\end{aligned}
\right.
\label{eq:programmatic_reward}
\end{equation}
where $\mathcal{V}$ denotes the set of structurally valid traces, $c(y)$ measures answer correctness, $a(y)$ is the action-valid rate, and $\Omega(y)$ collects repetition and answer-leakage penalties.
We set $\lambda=0.3$.
The reward evaluates both task success and protocol-consistent actions without a learned reward model.
See Appendix A.3 for reward details.

\noindent\textbf{Trajectory-level learning.}
For a group of complete responses $\{y^{(i)}\}_{i=1}^{G}$ sampled for the same input $x$, standard GRPO~\cite{shao2024deepseekmath} computes
\begin{equation}
A_x^{(i)}=\frac{R\!\left(x,y^{(i)}\right)-\bar{R}_x}{\sigma_{R,x}+\delta},
\label{eq:trajectory_advantage}
\end{equation}
where $\bar{R}_x$ and $\sigma_{R,x}$ denote the mean and standard deviation of the group rewards, respectively, and $\delta$ is a small constant for numerical stability.
The same advantage $A_x^{(i)}$ is applied to every generated token in $y^{(i)}$.
Although effective for optimizing response-level correctness and validity, trajectory-level GRPO ignores the event decomposition of Code-CoT and treats the structured response as a flat trajectory.
CE-GRPO retains the same programmatic reward and GRPO objective, but constructs comparison groups from shared intermediate states, as described next.

\subsection{Critical-Event Group Relative Policy Optimization}
\label{sec:method:cegrpo}

Steps~3--5 in~\cref{fig:overview} implement CE-GRPO through candidate-event selection, shared-prefix branching, and mixed policy optimization.

\noindent\textbf{Candidate-event selection.}
The Code-CoT tags deterministically segment each response into complete \texttt{think} and action events, eliminating the need for an auxiliary segmenter or step detector.
In Step~3, CE-GRPO combines a structural prior with type-normalized event entropy to select candidate events.
For an event $e$ with token set $\mathcal{T}_e$, let $\bar{H}(e)$ denote its mean token entropy.
Because entropy scales differ across event types, we normalize each event within its type:
\begin{equation}
\eta(e)
=
\frac{\bar{H}(e)-\mu_{\kappa(e)}}{\sigma_{\kappa(e)}},
\label{eq:type_normalized_entropy}
\end{equation}
where $\kappa(e)$ denotes the event type, and $\mu_{\kappa(e)}$ and $\sigma_{\kappa(e)}$ are its batch-level mean and standard deviation.
The structural prior provides semantically meaningful branch points, while $\eta(e)$ allocates the branching budget among comparable events.
The selector does not label an event as critical in advance; criticality is revealed only when sibling futures from the same prefix receive different terminal rewards.
See Appendix A.4 for selector details.

\noindent\textbf{Shared-prefix branching.}
In Step~4, for a candidate event $e_c$ beginning at position $s_c$, CE-GRPO retains the complete response prefix
\begin{equation}
z_c=y_{<s_c}.
\label{eq:shared_prefix}
\end{equation}

It fixes the image, question, and prefix $z_c$, then samples $G$ continuations through the final answer:
\begin{equation}
u^{(j)}
\sim
\pi_\theta(\,\cdot\mid x,z_c),
\qquad
y^{(j)}=z_c\oplus u^{(j)}.
\label{eq:shared_prefix_branching}
\end{equation}

Each branch is evaluated using the programmatic reward in~\cref{eq:programmatic_reward}, yielding the state-conditioned advantage
\begin{equation}
A_{z_c}^{(j)}
=
\frac{
R\!\left(x,z_c\oplus u^{(j)}\right)-\bar{R}_{z_c}
}{
\sigma_{R,z_c}+\delta
},
\label{eq:event_conditioned_advantage}
\end{equation}
where $\bar{R}_{z_c}$ and $\sigma_{R,z_c}$ are the mean and standard deviation of rewards among continuations sharing $z_c$.
Because the shared prefix belongs to the prompt and is excluded from the policy loss, the update applies only to the regenerated event and its downstream consequences.
If all continuations receive the same reward, the group contributes no policy update, so an uninformative selection primarily wastes computation rather than introducing spurious supervision.

\noindent\textbf{Mixed policy optimization.}
In Step~5, ordinary and shared-prefix prompts are mixed at a $1{:}1$ ratio.
Ordinary groups use the trajectory-level advantage in~\cref{eq:trajectory_advantage}, whereas shared-prefix groups use the event-conditioned advantage in~\cref{eq:event_conditioned_advantage}.
Both share the GRPO objective and programmatic reward, preserving global task learning while adding event-conditioned credit.
See Appendix A.4--A.5 for prefix and training details.

\section{Experiments}
\label{sec:exp}

\subsection{Experimental Setup}
\label{sec:setup}

\noindent\textbf{Benchmarks.}
We evaluate on nine geometry benchmarks spanning four categories: visual grounding with MathVerse~\citep{zhang2024mathverse}, VisOnlyQA-Syn and VisOnlyQA-Real~\citep{kamoi2025visonlyqa}, and MathVista-GPS~\citep{lu2023mathvista}; plane geometry with Geometry3K~\citep{lu2021intergps}, PGPS9K~\citep{zhang2023pgps}, and GeoQA~\citep{chen2021geoqa}; auxiliary construction with GeoLaux-mini~\citep{fu2026geolaux}; and process-level multimodal reasoning with MM-Math~\citep{sun2024mmmath}.
For compactness, result tables abbreviate MathVerse, VisOnlyQA-Syn, VisOnlyQA-Real, MathVista-GPS, Geometry3K, and GeoLaux-mini as MathV., VOQA-S, VOQA-R, GPS, Geo3K, and GeoLaux, respectively.

\noindent \textbf{Baselines.}
We compare with the native Qwen3-VL-8B-Instruct~\citep{bai2025qwen3vl}, prompting and SFT baselines, general post-training methods (DPO~\citep{rafailov2023dpo}, PPO~\citep{schulman2017ppo}, DAPO~\citep{yu2025dapo}, and trajectory-level GRPO~\citep{shao2024deepseekmath}), and critical-event or counterfactual methods (SRPO-style~\citep{samanta2026creditassignmentresetslanguage}, GPO~\citep{yu2025gpo}, CFPO~\citep{yu2026cfpo}, and GRPO-MA~\citep{wang2026branching}). External references include Qwen2.5-VL-7B-Instruct~\citep{bai2025qwen25vl}, VL-Rethinker-7B~\citep{wang2025vlrethinker}, MMR1-Math-v0-7B~\citep{mmr1team2025}, G-LLaVA-7B~\citep{gao2023gllava}, and the two-stage GDP-4B-RL~\citep{wang2026geoparsing} and GeoTikzBridge-8B~\citep{sun2026geotikzbridge} systems paired with Qwen3-VL-8B. 
External checkpoints are evaluated using their officially released prompt templates.

\noindent\textbf{Implementation.}
Code-CoT SFT initializes from Qwen3-VL-8B-Instruct, freezes the visual encoder, and updates the multimodal aligner and language model.
All in-house post-training methods start from this shared checkpoint; DPO and GPO use LoRA, while the remaining methods use full-parameter updates.
Protocol-constrained models must terminate with a non-empty \texttt{<answer>} block.
See Appendix A.5 and B.2--B.3 for implementation details.

\begin{table*}[!t]
\centering
\begin{minipage}{0.94\textwidth}
\centering
\scriptsize
\renewcommand{\arraystretch}{0.92}
\setlength{\tabcolsep}{2.6pt}
\setlength{\aboverulesep}{0.25ex}
\setlength{\belowrulesep}{0.35ex}
\setlength{\cmidrulesep}{0.35ex}
\resizebox{\linewidth}{!}{
\begin{tabular}{lcccccccccc}
\toprule[1pt]
\multicolumn{1}{c}{\multirow{2}{*}{\raisebox{-0.8ex}{\textbf{Method}}}} & \multicolumn{4}{c}{\textbf{Visual Grounding}} & \multicolumn{3}{c}{\textbf{Plane Geometry}} & \textbf{Aux. Constr.} & \textbf{Process-Level} & \multirow{2}{*}{\raisebox{-0.8ex}{\textbf{Avg.}}} \\
\cmidrule(lr){2-5}\cmidrule(lr){6-8}\cmidrule(lr){9-9}\cmidrule(lr){10-10}
\multicolumn{1}{c}{} & MathV. & VOQA-S & VOQA-R & GPS & Geo3K & PGPS9K & GeoQA & GeoLaux & MM-Math & \\
\midrule
\multicolumn{11}{c}{\textbf{\textit{Backbone: Qwen3-VL-8B-Instruct}}} \\
\midrule
\multicolumn{11}{l}{\textbf{\textit{Training-Free}}} \\
Qwen3-VL-8B-Instruct (native) & 56.85 & 51.13 & 63.73 & 83.19 & 64.52 & 59.60 & 83.16 & 77.27 & 72.10 & 67.95 \\
\quad with the Code-CoT prompt & 47.49 & 40.21 & 49.83 & 62.96 & 44.65 & 44.50 & 63.26 & 47.88 & 42.57 & 49.26 \\
\midrule
\multicolumn{11}{l}{\textbf{\textit{Supervised Fine-Tuning}}} \\
Code-CoT SFT & 55.08 & 62.89 & 70.85 & 81.48 & 53.31 & 52.10 & \underline{87.53} & \underline{84.24} & \underline{78.51} & 69.55 \\
\midrule
\multicolumn{11}{l}{\textbf{\textit{General Post-Training}}} \\
DPO (LoRA, $r{=}64$) & 49.06 & 60.62 & 70.85 & 73.15 & 46.69 & 49.10 & 76.66 & 65.15 & 67.77 & 62.12 \\
PPO (no KL) & 40.18 & 52.99 & 61.36 & 62.50 & 39.39 & 39.80 & 66.84 & 53.03 & 55.62 & 52.41 \\
DAPO & 55.99 & 58.56 & \textbf{75.93} & 76.85 & 51.44 & 54.40 & 84.08 & 73.03 & 74.50 & 67.20 \\
Trajectory-level GRPO & \underline{61.85} & \textbf{64.19} & 71.86 & \underline{86.11} & \textbf{70.29} & \underline{66.30} & 86.60 & 75.45 & 70.88 & \underline{72.61} \\
\midrule
\multicolumn{11}{l}{\textbf{\textit{Critical-Event and Counterfactual Post-Training}}} \\
SRPO-style (offline self-reset) & 56.19 & 62.47 & 72.88 & 82.41 & 55.52 & 54.60 & 86.21 & 77.58 & 75.90 & 69.31 \\
GPO (LoRA) & 46.73 & 58.35 & 68.81 & 67.59 & 45.33 & 45.60 & 73.47 & 54.55 & 64.86 & 58.37 \\
CFPO & 56.50 & 62.27 & 72.20 & 81.02 & 56.03 & 55.50 & 86.47 & 74.55 & 74.10 & 68.74 \\
GRPO-MA & 53.73 & 62.47 & 73.56 & 81.48 & 53.82 & 53.90 & 83.69 & 79.09 & 73.49 & 68.36 \\
\midrule
\rowcolor{gray!20}
\textbf{CE-GRPO (ours)} & \textbf{62.44}\ptdiff{+5.59} & \underline{63.47}\ptdiff{+12.34} & \underline{74.24}\ptdiff{+10.51} & \textbf{86.57}\ptdiff{+3.38} & \underline{67.40}\ptdiff{+2.88} & \textbf{66.90}\ptdiff{+7.30} & \textbf{92.44}\ptdiff{+9.28} & \textbf{90.61}\ptdiff{+13.34} & \textbf{80.32}\ptdiff{+8.22} & \textbf{76.04}\ptdiff{+8.09} \\
\bottomrule[1pt]
\end{tabular}}
\normalsize
\caption{Results on nine benchmarks; \textbf{Avg.} is the unweighted mean. Bold and underlined values are best and second best. CE-GRPO superscripts are gains over native Qwen3-VL-8B.}
\label{tab:main}
\end{minipage}

\vspace{0.1em}

\begin{minipage}{0.94\textwidth}
\centering
\scriptsize
\renewcommand{\arraystretch}{0.92}
\setlength{\tabcolsep}{2.6pt}
\setlength{\aboverulesep}{0.25ex}
\setlength{\belowrulesep}{0.35ex}
\setlength{\cmidrulesep}{0.35ex}
\resizebox{\linewidth}{!}{
\begin{tabular}{lcccccccccc}
\toprule[1pt]
\multicolumn{1}{c}{\multirow{2}{*}{\raisebox{-0.8ex}{\textbf{Method}}}} & \multicolumn{4}{c}{\textbf{Visual Grounding}} & \multicolumn{3}{c}{\textbf{Plane Geometry}} & \textbf{Aux. Constr.} & \textbf{Process-Level} & \multirow{2}{*}{\raisebox{-0.8ex}{\textbf{Avg.}}} \\
\cmidrule(lr){2-5}\cmidrule(lr){6-8}\cmidrule(lr){9-9}\cmidrule(lr){10-10}
\multicolumn{1}{c}{} & MathV. & VOQA-S & VOQA-R & GPS & Geo3K & PGPS9K & GeoQA & GeoLaux & MM-Math & \\
\midrule
\multicolumn{11}{l}{\textbf{\textit{External 7B/8B Checkpoints}}} \\
Qwen2.5-VL-7B-Instruct & 49.64 & 34.43 & 43.39 & 65.28 & 41.60 & 43.00 & 76.39 & 42.42 & 39.46 & 48.40 \\
VL-Rethinker-7B & 54.87 & 34.43 & 42.71 & 69.44 & 43.97 & 44.40 & 77.32 & 58.18 & 48.49 & 52.65 \\
MMR1-Math-v0-7B & 52.23 & 34.43 & 43.39 & 73.15 & 48.39 & 50.60 & 77.98 & 51.52 & 42.17 & 52.65 \\
G-LLaVA-7B & 20.13 & 25.98 & 28.14 & 39.81 & 8.83 & 8.60 & 61.27 & 3.03 & 11.14 & 22.99 \\
\midrule
\multicolumn{11}{l}{\textbf{\textit{Two-Stage Perception--Reasoning Systems}}} \\
GDP-4B-RL $\to$ Qwen3-VL-8B & \underline{61.65} & \underline{52.78} & 67.12 & \underline{83.80} & \textbf{71.82} & \textbf{74.20} & \underline{88.86} & \underline{77.88} & \underline{78.61} & \underline{72.97} \\
GeoTikzBridge-8B $\to$ Qwen3-VL-8B & 60.13 & 48.87 & \underline{69.83} & 80.09 & 62.48 & 64.10 & 87.93 & 74.55 & 78.21 & 69.58 \\
\midrule
\rowcolor{gray!20}
\textbf{CE-GRPO (ours)} & \textbf{62.44}\ptdiff{+0.79} & \textbf{63.47}\ptdiff{+10.69} & \textbf{74.24}\ptdiff{+7.12} & \textbf{86.57}\ptdiff{+2.77} & \underline{67.40}\ptdiff{-4.42} & \underline{66.90}\ptdiff{-7.30} & \textbf{92.44}\ptdiff{+3.58} & \textbf{90.61}\ptdiff{+12.73} & \textbf{80.32}\ptdiff{+1.71} & \textbf{76.04}\ptdiff{+3.07} \\
\bottomrule[1pt]
\end{tabular}}
\normalsize
\caption{External-checkpoint and two-stage results; CE-GRPO is repeated for reference. Bold and underlined values are best and second best. Its superscripts are gains over GDP-4B-RL $\to$ Qwen3-VL-8B; two-stage systems use an extra call.}
\label{tab:external}
\end{minipage}
\end{table*}

\subsection{Main Results}
\label{sec:main}

\Cref{tab:main,tab:external} report the main results.

\noindent\textbf{Overall performance.}
Code-CoT prompting alone obtains $49.26$, $18.69$ points below the native backbone, while SFT raises the average to $69.55$, showing that the protocol must be learned rather than prompted.
CE-GRPO achieves the best average of $76.04$, outperforming the backbone, Code-CoT SFT, trajectory-level GRPO, and the strongest fine-grained baseline by $8.09$, $6.49$, $3.43$, and $6.73$ points, respectively.
It also surpasses the backbone on all nine benchmarks.

\noindent\textbf{Gains on dependency-heavy reasoning.}
Relative to trajectory-level GRPO, CE-GRPO gains $15.16$ points on GeoLaux-mini and $9.44$ on MM-Math, where intermediate constructions and decisions affect multiple later steps.
By contrast, gains are smaller or mixed on visual-grounding and standard plane-geometry tasks.
This pattern supports assigning credit to outcome-sensitive intermediate events rather than uniformly across the full trajectory.

\noindent\textbf{Comparison with two-stage systems.}
With a single model call, CE-GRPO surpasses GDP-4B-RL $\to$ Qwen3-VL-8B by $3.07$ points on average, winning seven of nine benchmarks, and outperforms GeoTikzBridge-8B $\to$ Qwen3-VL-8B on all nine.
GDP leads only on Geometry3K and PGPS9K, suggesting that fixed symbolic parsing favors relation-centric geometry, whereas integrated perception and reasoning better handle dynamic constructions and visual evidence reuse.

\subsection{Ablation Study}
\label{sec:ablation}

\noindent\textbf{Setup.}
\Cref{tab:selector} compares event selectors with all other components fixed, reporting each variant's best validation checkpoint.
\emph{Random prefix} samples events uniformly; \emph{entropy only} ranks them by type-normalized mean token entropy; \emph{structure only} uses the first geometry-grounded \texttt{<think>} event and \texttt{<reference>} action; and \emph{structure + entropy} ranks these structurally anchored candidates by entropy.

\noindent\textbf{Structure localizes useful events; entropy prioritizes them.}
Entropy alone provides only a $0.35$-point gain over random selection, whereas structure raises the average to $74.26$ and reduces the unclosed-response rate from $12.31\%$ to $7.07\%$.
Combining both signals performs best, reaching $76.04$ with the lowest unclosed rate of $4.73\%$, and is best or tied on seven of nine benchmarks.
Its gains over structure alone are particularly pronounced on Geometry3K ($+5.43$) and GeoLaux-mini ($+3.34$), indicating that structural priors identify semantically meaningful boundaries while entropy distinguishes informative candidates among them.

\section{Discussion}
\label{sec:discussion}

\begin{table}[t]
\centering
\small
\setlength{\tabcolsep}{4pt}
\begin{tabular}{lccc}
\toprule
Condition & Macro recall & Render succ. & Recall $|$ rendered \\
\midrule
Base + prompt
& 55.21 & 89.0 & 62.0 \\
Code-CoT SFT
& 70.16 & 95.0 & 73.9 \\
CE-GRPO
& \textbf{80.43} & \textbf{99.0} & \textbf{81.2} \\
\bottomrule
\end{tabular}
\caption{Diagram-to-code fidelity on 100 MathVerse-TD problems. All values are percentages; failed renders receive zero macro recall.}
\label{tab:code-fidelity}
\end{table}

\noindent\textbf{Code-CoT training improves diagram-to-code fidelity.}
To examine whether training improves the executable visual representation itself, \cref{tab:code-fidelity} evaluates the first \texttt{<perception>} block on $100$ MathVerse-TD problems.
For each diagram, Gemini-3.1-Pro-Preview extracts a fixed set of visible geometric facts, and a condition-blind judge scores each fact as fully covered ($1$), partially covered ($0.5$), or missing or contradicted ($0$); failed renders receive zero recall.
Code-CoT SFT raises macro recall from $55.21\%$ to $70.16\%$, while CE-GRPO further improves it to $80.43\%$.
Render success increases from $89.0\%$ to $99.0\%$, and recall on successful renders from $62.0\%$ to $81.2\%$, showing gains in both code executability and geometric fidelity.

\begin{table*}[t]
\centering
\scriptsize
\setlength{\tabcolsep}{2.4pt}
\begin{tabular*}{\textwidth}{@{\extracolsep{\fill}}lccccccccccc@{}}
\toprule
\multicolumn{1}{c}{\multirow{2}{*}{\raisebox{-0.8ex}{\textbf{Selection signal}}}}
& \multicolumn{4}{c}{\textbf{Visual Grounding}}
& \multicolumn{3}{c}{\textbf{Plane Geometry}}
& \textbf{Aux. Constr.}
& \textbf{Process-Level}
& \multirow{2}{*}{\raisebox{-0.8ex}{\textbf{Avg.}}}
& \multirow{2}{*}{\raisebox{-0.8ex}{\textbf{Uncl.}}} \\
\cmidrule(lr){2-5}
\cmidrule(lr){6-8}
\cmidrule(lr){9-9}
\cmidrule(lr){10-10}
\multicolumn{1}{c}{}
& MathV. & VOQA-S & VOQA-R & GPS
& Geo3K & PGPS9K & GeoQA
& GeoLaux & MM-Math & & \\
\midrule
Random prefix
& 59.42
& \textbf{65.57}
& \textbf{75.25}
& \underline{85.65}
& 61.46
& 60.00
& 89.92
& 78.48
& 76.61
& 72.48
& 12.31\% \\

Entropy only
& 58.40
& 64.33
& 72.88
& \underline{85.65}
& 60.78
& 61.50
& \underline{90.98}
& 82.42
& 78.51
& 72.83
& 10.26\% \\

Structure only
& \underline{61.19}
& \underline{65.15}
& 72.20
& 83.80
& \underline{61.97}
& \underline{64.10}
& \textbf{92.44}
& \underline{87.27}
& \underline{80.22}
& \underline{74.26}
& \underline{7.07\%} \\

\textbf{Structure + entropy}
& \textbf{62.44}
& 63.47
& \underline{74.24}
& \textbf{86.57}
& \textbf{67.40}
& \textbf{66.90}
& \textbf{92.44}
& \textbf{90.61}
& \textbf{80.32}
& \textbf{76.04}
& \textbf{4.73\%} \\
\bottomrule
\end{tabular*}
\caption{Event-selector ablation. \textbf{Uncl.} is the mean invalid-termination
rate. Bold and underlined values are best and second best among all selectors;
lower \textbf{Uncl.} is better.}
\label{tab:selector}
\end{table*}

\begin{table}[t]
\centering
\scriptsize
\begin{tabular*}{\linewidth}{@{\extracolsep{\fill}}lccccc@{}}
\toprule
\multirow{2}{*}{\textbf{Benchmark}}
& \multicolumn{2}{c}{\textbf{Traj.\ GRPO}}
& \multicolumn{2}{c}{\textbf{CE-GRPO}}
& \multirow{2}{*}{\(\boldsymbol{\Delta}\)} \\
\cmidrule(lr){2-3}
\cmidrule(lr){4-5}
& \makecell{Closed Acc.}
& \makecell{Uncl.}
& \makecell{Closed Acc.}
& \makecell{Uncl.}
& \\
\midrule
MathV.
& 64.20 & 3.66
& 66.15 & 5.61
& $+1.95$ \\

VOQA-S
& 64.19 & 0.00
& 63.73 & 0.41
& $-0.46$ \\

VOQA-R
& 71.86 & 0.00
& 74.24 & 0.00
& $+2.38$ \\

GPS
& 89.42 & 3.70
& 91.66 & 5.56
& $+2.24$ \\

Geo3K
& 75.27 & 6.62
& 73.65 & 8.49
& $-1.62$ \\

PGPS9K
& 70.61 & 6.10
& 75.68 & 11.60
& $+5.07$ \\

GeoQA
& 90.32 & 4.12
& 94.96 & 2.65
& $+4.64$ \\

GeoLaux
& 83.00 & 9.10
& 95.53 & 5.15
& $+12.53$ \\

MM-Math
& 74.39 & 4.72
& 82.81 & 3.01
& $+8.42$ \\ \midrule
\textbf{Mean}
& \textbf{75.92}
& \textbf{4.22}
& \textbf{79.82}
& \textbf{4.72}
& $\mathbf{+3.91}$ \\
\bottomrule
\end{tabular*}
\caption{
Closed-only accuracy and unclosed-response rate (\%).
\(\Delta\) denotes the closed-only accuracy difference between CE-GRPO and trajectory-level GRPO.
}
\label{tab:closedonly}
\end{table}

\noindent\textbf{CE-GRPO Improves Solution Quality on Valid Outputs.} Strict accuracy combines solution correctness with valid termination. We therefore report each model's unclosed rate and recompute accuracy on valid outputs, as shown in~\cref{tab:closedonly}. CE-GRPO outperforms trajectory-level GRPO on seven of nine benchmarks, with a mean gain of $3.91$ points. The largest gains occur on GeoLaux-mini ($+12.53$), MM-Math ($+8.42$), and PGPS9K ($+5.07$). These results show that CE-GRPO improves auxiliary construction and process-level reasoning within completed solutions, not only response completion.

\begin{table}[t]
\centering
\scriptsize
\setlength{\tabcolsep}{1.3pt}
\renewcommand{\arraystretch}{1.05}
\begin{tabular}{lrrrrrrrr}
\toprule
\textbf{Model} & \textbf{TD} & \textbf{TL} & \textbf{VI} & \textbf{VD} & \textbf{VO} & \textbf{All} & \textbf{Gap}$\downarrow$ & \textbf{SD}$\downarrow$ \\
\midrule
Backbone
& 69.92 & 62.31 & 55.96 & 56.22 & 39.85 & 56.85 & 30.07 & 9.91 \\
\rowcolor{gray!15}
\textbf{CE-GRPO}
& 69.04 & 65.99 & 62.31 & 59.90 & 54.95 & 62.44 & \textbf{14.09} & \textbf{4.87} \\
\midrule
$\Delta$
& $-0.88$ & $+3.68$ & $+6.35$ & $+3.68$ & $+15.10$ & $+5.59$ & $\mathbf{-15.98}$ & $\mathbf{-5.03}$ \\
\bottomrule
\end{tabular}
\caption{
Accuracy (\%) across five MathVerse variants, from Text Dominant (TD) to Vision Only (VO). All is the mean; Gap is $|\mathrm{TD}-\mathrm{VO}|$; SD is the population standard deviation ($\downarrow$: better). $\Delta$ is CE-GRPO minus Backbone.
}
\label{tab:mathverse-subsets}
\vspace{-0.5em}
\end{table}

\noindent\textbf{Code-grounded training narrows the modality gap.}
MathVerse shifts the same $788$ problems from text-dominant to vision-only inputs.
CE-GRPO leaves TD nearly unchanged ($-0.88$) but gains $15.10$ points on VO, reducing the TD--VO gap from $30.07$ to $14.09$ and SD from $9.91$ to $4.87$---roughly halving both.
The concentration of gains on vision-dependent variants suggests that Code-CoT better preserves and reuses diagram-derived relations, rather than merely improving text-dominant reasoning, thereby mitigating the perception bottleneck in~\cref{tab:prestudy}.

\begin{figure}[t]
\centering
\includegraphics[width=\linewidth]{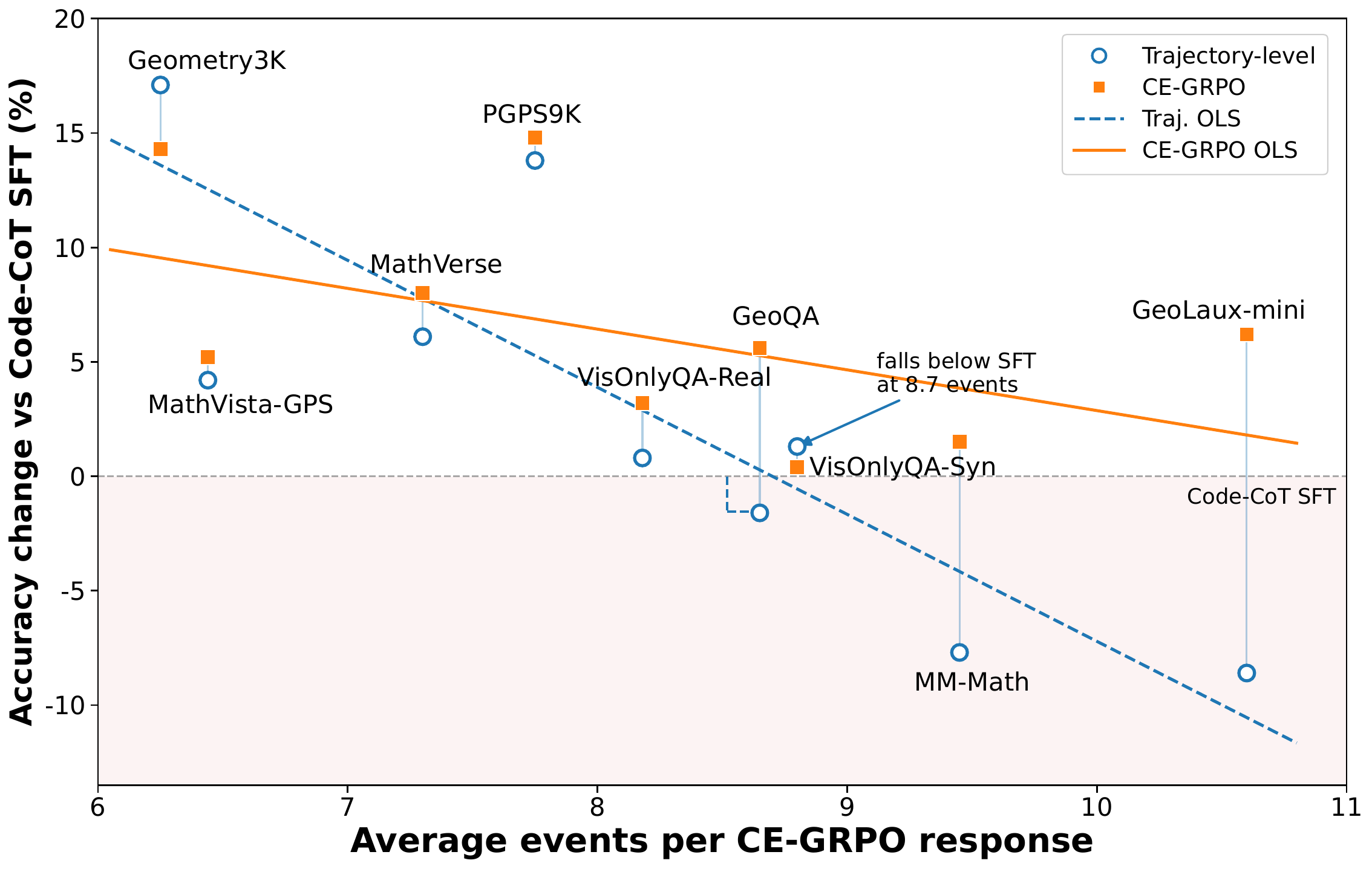}
\caption{
Accuracy change relative to Code-CoT SFT versus the mean number of complete
\texttt{\textless think\textgreater}/\texttt{\textless action\textgreater}
events per response.
Gray segments connect matched benchmarks; lines show OLS fits, and the
horizontal dashed line marks the SFT baseline.
}
\label{fig:event-count}
\end{figure}

\noindent\textbf{CE-GRPO's advantage widens with event count.}
Across the nine benchmarks, we relate the mean number of complete \texttt{<think>} and \texttt{<action>} events in CE-GRPO outputs to each method's gain over the same Code-CoT SFT checkpoint.
As shown in~\cref{fig:event-count}, CE-GRPO's margin over trajectory-level GRPO increases by $3.77$ points per additional event ($r=0.866$, exact $p=0.0016$), remaining significant under leave-one-benchmark-out tests ($p\leq0.028$).
This widening is driven by trajectory-level GRPO, whose gain over SFT declines sharply with event count ($-5.55$ points/event, $p=0.0062$), whereas the corresponding CE-GRPO trend is weaker and nonsignificant ($-1.78$ points/event, $p=0.2048$).
This pattern suggests that whole-trajectory credit diffuses over longer reasoning chains, whereas event-localized credit remains effective. For further details, see Appendix C.2.

\section{Conclusion}
\label{sec:conclusion}

This work introduces \emph{credit-addressable reasoning}, a representation--optimization principle in which the semantic units exposed during inference also define where learning compares alternatives and assigns credit.
\textbf{Code-CoT} realizes this principle by retaining the original diagram, representing geometric relations as line-addressable executable code, and organizing reasoning into typed events, while \textbf{CE-GRPO} branches complete continuations from shared event prefixes and converts terminal outcome variation into event-conditioned advantages.
Across nine geometry benchmarks, CE-GRPO achieves an average accuracy of $76.04$, outperforming Qwen3-VL-8B and trajectory-level GRPO by $8.09$ and $3.43$ points, respectively, and retaining a $3.91$-point advantage on validly terminated responses.
Its relative advantage further increases with the number of intermediate events, while selector analyses confirm the value of structurally meaningful event boundaries.
The framework is broadly applicable to reasoning tasks with explicit intermediate structure and delayed outcome supervision.
These results demonstrate that aligning reasoning representations with credit assignment provides an effective direction for long, dependency-heavy multimodal reasoning.

\bibliography{aaai2027}

\clearpage

\appendix
\section*{Appendix}

\section{Methodological Details}
\label{app:details}

\subsection{Code-CoT Protocol and Verification}
\label{app:protocol}

\noindent\textbf{Response protocol.}
Code-CoT places a line-addressable diagram program inside the reasoning trace and treats it as a persistent representation throughout the solution.
Given an image $I$ and question $Q$, the model produces a response following the protocol below:
\begin{center}
{\setlength{\fboxsep}{6pt}%
\fcolorbox{black!25}{white}{%
\begin{minipage}{0.88\columnwidth}
\small
\textcolor{blue!70!black}{\texttt{<perception>}}\newline
\hspace*{1em}\textit{line-numbered executable Matplotlib code}\newline
\textcolor{blue!70!black}{\texttt{</perception>}}\par
\hfill$\downarrow$\quad\textit{grounds subsequent reasoning}\quad$\downarrow$\hfill\mbox{}\par
\textcolor{purple!80!black}{\texttt{PLAN:}}\ \textit{one-line solution route}\par
\hfill$\downarrow$\hfill\mbox{}\par
\textcolor{teal!75!black}{\texttt{<think>}}\ \textit{reasoning step}\ \textcolor{teal!75!black}{\texttt{</think>}}\par
\hfill$\downarrow$\quad\textit{when a code operation is needed}\quad$\downarrow$\hfill\mbox{}\par
\textcolor{orange!85!black}{\texttt{<action type="...">}}\newline
\hspace*{1em}\textit{referenced lines or executable code}\newline
\textcolor{orange!85!black}{\texttt{</action>}}\par
\hfill$\downarrow$\quad\textit{use the result in the next step}\quad$\downarrow$\hfill\mbox{}\par
\textcolor{teal!75!black}{\texttt{<think>}}\ \textit{next reasoning step}\ \textcolor{teal!75!black}{\texttt{</think>}}\par
\hfill$\vdots$\quad\textit{repeat as needed}\quad$\vdots$\hfill\mbox{}\par
\textcolor{green!45!black}{\texttt{<answer>}}\ \textit{final answer}\ \textcolor{green!45!black}{\texttt{</answer>}}
\end{minipage}}}
\end{center}

A valid response begins with exactly one \texttt{<perception>} block, followed by a one-line plan prefixed with \texttt{PLAN:}, an ordered sequence of reasoning and action events, and exactly one non-empty \texttt{<answer>} block.
The perception block is generated once and remains unchanged throughout the trace.
Each intermediate event is either a reasoning block enclosed by \texttt{<think>} and \texttt{</think>} or an action block of the form
\texttt{<action type="...">}\ldots\texttt{</action>}.
Actions are inserted only when the solution requires a code-grounded retrieval or intervention, rather than after every reasoning step.

\noindent\textbf{Typed actions.}
Code-CoT defines three action types with distinct effects on the working state.
Their operational roles and validity conditions are summarized in~\cref{tab:protocol_actions}.

A \textsf{reference} action does not modify the working figure; it exposes the perception lines grounding a visual fact.
An \textsf{auxiliary} action may introduce connections or extensions, perpendicular or parallel lines, points or midpoints, angle bisectors, circles, transformations, variables, or other explicit constructions.
A \textsf{coordinate} action changes the representation of the working geometry by introducing a code-based frame and concrete coordinates.
The outputs of auxiliary and coordinate actions become part of the state available to subsequent \texttt{think} events.

\noindent\textbf{Programmatic verification.}
We parse each response into its perception code, plan, intermediate events, and final answer before evaluating its contents.
The structural checker requires exactly one perception block, one plan, and one non-empty answer; balanced and correctly ordered tags; recognized action types; and at least two action events.
Responses with missing, duplicated, incomplete, or improperly nested protocol blocks are structurally invalid.

For structurally valid responses, the executability checker runs the perception program and executable action contents in their generated order.
The grounding checker validates every \textsf{reference} action against the line addresses and contents of the original perception block.
Type-specific checkers then apply the conditions in~\cref{tab:protocol_actions} to the auxiliary and coordinate actions.
Answer correctness and additional behavioral penalties are evaluated separately as part of the programmatic reward in~\cref{app:optimization}.

For a response $Y$, let $n(Y)$ denote its number of action events and $v(Y)$ the number satisfying their type-specific validity conditions.
We define the action-validity rate as
\begin{equation}
r_{\mathrm{act}}(Y)=
\frac{v(Y)}{n(Y)}.
\label{eq:action_validity}
\end{equation}

Since a structurally valid response contains at least two actions, the denominator is always nonzero.
This rate provides a graded signal beyond terminal answer correctness, rewarding traces whose code operations are individually well formed and grounded.

\begin{table}[!t]
\centering
\footnotesize
\setlength{\tabcolsep}{3pt}
\begin{tabular}{
p{0.18\columnwidth}
p{0.32\columnwidth}
p{0.45\columnwidth}}
\toprule
\textbf{Type} &
\textbf{Role} &
\textbf{Validity conditions} \\
\midrule

\textsf{reference}
&
Retrieves perception lines supporting a diagram fact before its first use.
&
Every cited line exists in the perception block, the copied content matches the corresponding line, and at most eight lines are included.
\\[2pt]

\textsf{auxiliary}
&
Adds an object or relation to extend the working geometry.
&
The operation is supported, all referenced objects already exist, and concrete executable code is provided for every new object.
\\[2pt]

\textsf{coordinate}
&
Establishes an executable coordinate frame for subsequent reasoning.
&
The action calls \texttt{set\_frame}, contains no ellipsis placeholder, and provides concrete coordinate assignments.
\\
\bottomrule
\end{tabular}
\caption{Operational roles and validity conditions of Code-CoT actions.}
\label{tab:protocol_actions}
\end{table}

\noindent\textbf{Deterministic event parsing.}
The protocol tags also provide deterministic boundaries for event-level optimization.
A single-pass parser retains each complete \texttt{think} or action block as one event and records its type and span without a learned segmenter or heuristic step detector.
The perception block, plan, final answer, and untagged text are excluded from the candidate-event pool.
Consequently, every candidate considered by CE-GRPO corresponds to a complete protocol unit, and its preceding prefix can be recovered exactly.

\subsection{Training Data Construction}
\label{app:data}

\noindent\textbf{Supervised fine-tuning (SFT) data.}
We construct the SFT set from eight geometry datasets covering synthetic diagrams, formal geometry, auxiliary constructions, and multimodal mathematical reasoning.
We remove question- and image-level overlaps with the evaluation sets before trace synthesis.
After filtering, the SFT set contains $18,302$ examples, with the source distribution reported in~\cref{tab:sft-sources}.

\begin{table}[!t]
\centering
\small
\begin{tabular}{lr}
\toprule
\textbf{Source dataset} & \textbf{Samples} \\
\midrule
MultiMath-Geo~\cite{peng2024multimath}  & 4,974 \\
FormalGeo7K~\cite{zhang2024formalgeo}     & 1,931 \\
UniGeo~\cite{chen2022unigeo}          & 2,313 \\
GeoAux~\cite{fu2026geolaux}          & 240 \\
Geo170K~\cite{gao2025gllava}        & 816 \\
GeoSym127K~\cite{jing2026geosym127k}     & 1,183 \\
MultiMath-300K~\cite{peng2024multimath}  & 4,227 \\
PGPS9K~\cite{zhang2023pgps}          & 2,618 \\
\midrule
\textbf{Total} & \textbf{18,302} \\
\bottomrule
\end{tabular}
\caption{Retained SFT samples from each source dataset.}
\label{tab:sft-sources}
\end{table}

\noindent\textbf{Code-CoT trace synthesis.}
For each retained image $I$, Gemini-3.1-Pro~\cite{googledeepmind2026gemini31pro} transcribes the diagram into line-numbered, executable Matplotlib code $C$.
DeepSeek-V4-Pro~\citep{deepseekv4pro} then receives the question $Q$ and code $C$ and generates the one-line plan $P$, interleaved reasoning and action events, and final answer $F$.
The reference answer is available only for silent consistency checking during synthesis.

Each synthesized example defines the structured target
\begin{equation}
(I,Q)\longmapsto Y=\big[C;,P;,(e_\ell)*{\ell=1}^{L};,F\big],\quad
e*\ell\in\mathcal{E},
\label{eq:sft-target}
\end{equation}
where $\mathcal{E}$ contains the \texttt{think}, \texttt{reference}, \texttt{auxiliary}, and \texttt{coordinate} event types defined in~\cref{app:protocol}.
The student model receives the original image and question and predicts the complete response $Y$, including both the perception program and the code-grounded reasoning trace.

We retain a synthesized trace only if it passes the protocol checks described in~\cref{app:protocol}.
Specifically, the response must have a valid structure, executable perception and action code, line-grounded reference actions, valid typed actions, and a correct final answer.
This filtering ensures that SFT supervision provides both a valid Code-CoT format and an executable reasoning trajectory.

\noindent\textbf{Reinforcement learning (RL) problem pool.}
The RL set contains original geometry problems rather than pre-synthesized Code-CoT traces.
The current policy must therefore generate the perception program, reasoning events, actions, and final answer during each rollout.
We emphasize harder problems from the hard and expert tiers of GeoSym127K~\cite{jing2026geosym127k} and unused PGPS9K~\cite{zhang2023pgps} training samples, while including a smaller GEOQA~\cite{chen2021geoqa} subset to broaden coverage.
After deduplication against the SFT and evaluation sets, the RL pool contains $11,450$ problems.
Counterfactual prefixes derived from this pool are constructed separately through the candidate-selection and prefix-harvesting procedure.

\subsection{Optimization Details}
\label{app:optimization}

We optimize Code-CoT in two stages.
First, supervised fine-tuning learns the complete structured response defined in~\cref{eq:codecot_output}.
The resulting checkpoint then initializes reinforcement learning, where ordinary and shared-prefix prompts use the same programmatic reward and clipped GRPO objective but different group-relative advantages.

\noindent\textbf{Supervised objective.}
Given an input $x=(I,Q)$ and its target response $Y=(Y_1,\ldots,Y_{|Y|})$, we minimize the autoregressive negative log-likelihood
\begin{equation}
\mathcal{L}_{\mathrm{SFT}}(\theta)=
-\mathbb{E}_{(x,Y)\sim\mathcal{D}_{\mathrm{SFT}}}
\left[
\frac{1}{|Y|}
\sum_{t=1}^{|Y|}
\log \pi_\theta\!\left(Y_t\mid x,Y_{<t}\right)
\right].
\label{eq:sft_objective}
\end{equation}

The loss covers the perception code, solution plan, typed events, and final answer.

\noindent\textbf{Programmatic reward.}
For a generated response $y$, we define
\begin{equation}
R(x,y)
=
\left\{
\begin{aligned}
-1,
& \quad y\notin\mathcal{V},\\
\operatorname{clip}\!\left(q(y),-1,1.3\right),
& \quad y\in\mathcal{V},
\end{aligned}
\right.
\label{eq:app_programmatic_reward}
\end{equation}
where $\mathcal{V}$ is the set of structurally valid responses and
\begin{equation}
q(y)=c(y)+0.3\,r_{\mathrm{act}}(y)-\Omega(y).
\label{eq:valid_response_score}
\end{equation}

Here, $c(y)=\mathbb{I}\!\left[\hat{a}(y)\equiv a^\star\right]$ denotes answer correctness, and $r_{\mathrm{act}}(y)$ is the action-validity rate in~\cref{eq:action_validity}.
The behavioral penalty is
\begin{equation}
\Omega(y)=\Omega_{\mathrm{dup}}(y)+\Omega_{\mathrm{rep}}(y)+\Omega_{\mathrm{leak}}(y),
\label{eq:behavioral_penalty}
\end{equation}
where the three terms penalize duplicated actions, repetitive generation, and answer leakage, respectively.
Multiple-choice answers use exact option matching, while numerical answers allow a $1\%$ relative tolerance.
The reward reaches $1.3$ when the answer is correct, all actions are valid, and no behavioral penalty is triggered.

\noindent\textbf{Group-relative policy objective.}
Let $\xi$ denote either an ordinary problem prompt $x$ or a shared-prefix prompt $(x,z_c)$, and let $g^{(i)}$ be the generated sequence for the $i$-th member of a group.
For ordinary prompts, $g^{(i)}$ is the complete response and uses the trajectory-level advantage in~\cref{eq:trajectory_advantage}.
For shared-prefix prompts, $g^{(i)}$ is the regenerated continuation and uses the event-conditioned advantage in~\cref{eq:event_conditioned_advantage}.
For generated token $g_t^{(i)}$, the policy ratio is
\begin{equation}
\rho_{i,t}(\theta)=\frac{
\pi_\theta\!\left(g_t^{(i)}\mid \xi,g_{<t}^{(i)}\right)
}{
\pi_{\theta_{\mathrm{old}}}\!\left(g_t^{(i)}\mid \xi,g_{<t}^{(i)}\right)
}.
\label{eq:policy_ratio}
\end{equation}

Both prompt types maximize
\begin{equation}
\begin{aligned}
J_{\mathrm{GRPO}}(\theta)
={}&
\mathbb{E}\Bigg[
\frac{1}{Z}
\sum_{i=1}^{G}
\sum_{t=1}^{|g^{(i)}|}
\min\Bigg(
\rho_{i,t}(\theta) A_{\xi}^{(i)}, \\
&\qquad
\operatorname{clip}\!\Big(
\rho_{i,t}(\theta),
1-\epsilon_{\mathrm{lo}},
1+\epsilon_{\mathrm{hi}}
\Big)
A_{\xi}^{(i)}
\Bigg)
\Bigg].
\end{aligned}
\label{eq:grpo_objective}
\end{equation}
where $Z=\sum_{i=1}^{G}|g^{(i)}|$ is the total number of generated tokens in the group.
We use $\delta=10^{-6}$ in the advantage normalization, $\epsilon_{\mathrm{lo}}=0.2$, and $\epsilon_{\mathrm{hi}}=0.28$, without an additional KL penalty.
Therefore, CE-GRPO changes the prompt construction and grouping state while retaining the same reward and clipped policy objective.
Candidate selection and prefix construction are detailed in~\cref{app:selection}, and the complete training procedure is provided in~\cref{app:procedure}.

\subsection{Candidate-Event Selection and Prefix Construction}
\label{app:selection}

\noindent\textbf{Event parsing and entropy estimation.}
We parse each Code-CoT response using the deterministic protocol boundaries defined in~\cref{app:protocol}.
The candidate pool contains complete \texttt{think}, \texttt{reference}, \texttt{auxiliary}, and \texttt{coordinate} events after the perception block.
The perception block, solution plan, final answer, and untagged text are excluded.
For token position $t$, we approximate its entropy using the recorded top-$20$ vocabulary candidates:
\begin{equation}
H_t
=
-\sum_{v\in\mathcal{V}_{20}(t)}
p_{t,v}\log p_{t,v},
\label{eq:top20_entropy}
\end{equation}
where $\mathcal{V}_{20}(t)$ denotes the top-$20$ candidates and $p_{t,v}$ is the probability assigned to candidate $v$.
For an event $e$ with token set $\mathcal{T}_e$, we compute
\begin{equation}
\bar{H}(e)
=
\frac{1}{|\mathcal{T}_e|}
\sum_{t\in\mathcal{T}_e} H_t.
\label{eq:event_entropy}
\end{equation}

We then standardize $\bar{H}(e)$ within its event type using~\cref{eq:type_normalized_entropy}.
This normalization is necessary because mean raw entropy differs substantially across event types; in our harvested traces, the mean entropy of \texttt{think} events is approximately $7.7\times$ that of \texttt{reference} events.

\noindent\textbf{Candidate-selection rule.}
The selector combines a structural candidate with an entropy-based candidate.
The structural candidate is initialized as the first complete \texttt{think} event, prioritizing the earliest explicit reasoning decision after perception.
Let $e_{\mathrm{first}}$ denote this event and $e_{\mathrm{think}}^\star$ the \texttt{think} event with the highest type-normalized entropy.
We replace $e_{\mathrm{first}}$ with $e_{\mathrm{think}}^\star$ only when
\begin{equation}
\eta\!\left(e_{\mathrm{think}}^\star\right)
-
\eta\!\left(e_{\mathrm{first}}\right)
>
1.
\label{eq:think_replacement}
\end{equation}
The second candidate is the action event with the highest type-normalized entropy.
With probability $0.15$, one selected candidate is replaced by an event sampled uniformly from the candidate pool to preserve exploratory coverage.
Offline selector evaluation uses both ordered candidates, whereas CE-GRPO training uses only the first returned candidate.
The selector therefore proposes semantically valid branch points but does not label them as critical in advance.

\noindent\textbf{Prefix construction.}
We run the Code-CoT SFT policy on $2{,}000$ problems from the RL pool and sample two complete trajectories per problem using the same stochastic decoding configuration adopted for RL rollout collection.
For the selected event $e_c$ beginning at position $s_c$, we retain the complete prefix $z_c=y_{<s_c}$ defined in~\cref{eq:shared_prefix}.
This truncation removes the candidate event and its entire suffix, requiring the policy to regenerate the event and continue through the final answer.
Length and protocol-structure filters yield $3{,}270$ unique shared prefixes.
The prefixes are collected once from the initial SFT policy and remain fixed throughout reinforcement learning.

\noindent\textbf{Counterfactual-prefix training rows.}
Each retained prefix is paired with its original image, question, and reference answer.
During rollout, the prefix is supplied as an assistant prefill, so it conditions generation but is excluded from the policy loss.
Each prefix is duplicated at most four times, producing $11{,}450$ counterfactual-prefix rows.
These rows are mixed at a $1{:}1$ ratio with the $11{,}450$ ordinary RL problems.
Ordinary rows construct groups from the original problem and use~\cref{eq:trajectory_advantage}, whereas counterfactual-prefix rows construct groups from a shared intermediate state and use~\cref{eq:event_conditioned_advantage}.
If all continuations from a prefix receive the same reward, the group has zero variance and contributes no policy update.

\begin{table*}[!t]
\centering
\footnotesize
\setlength{\tabcolsep}{5pt}
\renewcommand{\arraystretch}{1.12}
\begin{tabularx}{\textwidth}{
p{0.13\textwidth}
p{0.18\textwidth}
Y
p{0.30\textwidth}}
\toprule
\textbf{Template}
&
\textbf{Input}
&
\textbf{Instruction and expected output}
&
\textbf{Key constraints}
\\
\midrule

\textbf{Diagram transcription}
&
Image $I$
&
Convert the diagram into line-numbered, executable Matplotlib code that records visible geometric objects, labels, measurements, and relations.
Output only the perception program.
&
Preserve the diagram structure; use explicit executable statements; include no ellipses or unresolved placeholders.
\\[2pt]

\textbf{Trace synthesis}
&
Question $Q$, perception code $C$, and reference answer $a^\star$
&
Generate a one-line plan, an interleaved sequence of
\texttt{think} and typed \texttt{action} events, and one non-empty
\texttt{answer} block.
&
Ground diagram facts with \texttt{reference}; use
\texttt{auxiliary} and \texttt{coordinate} only when needed.
The reference answer is used only for silent consistency checking.
\\[2pt]

\textbf{Code-CoT policy}
&
Image $I$ and question $Q$
&
Generate the complete Code-CoT response:
a \texttt{perception} block, a plan, typed reasoning events, and the final answer.
&
Retain the original image during reasoning; generate concrete executable actions; invoke no external geometry solver.
\\[2pt]

\textbf{Ordinary RL}
&
Code-CoT policy instruction, image $I$, and question $Q$
&
Generate the complete response from the beginning of the
\texttt{perception} block through the final answer.
&
All generated tokens participate in the policy loss.
The response is evaluated using~\cref{eq:programmatic_reward}.
\\[2pt]

\textbf{Shared-prefix RL}
&
Code-CoT policy instruction, image $I$, question $Q$, and assistant prefix $z_c$
&
Continue generation from the boundary immediately preceding a candidate event, regenerating that event and its complete suffix.
&
The prefix conditions generation but is excluded from the policy loss.
Only the continuation receives the event-conditioned advantage.
\\

\bottomrule
\end{tabularx}
\caption{
Prompt templates used for data synthesis, supervised learning, and reinforcement learning.
Ordinary and shared-prefix RL use the same policy instruction; the latter additionally supplies the Code-CoT prefix preceding a candidate event as an assistant prefill.
}
\label{tab:prompt_templates}
\end{table*}

\begin{table*}[!t]
\centering
\small
\setlength{\tabcolsep}{5pt}
\renewcommand{\arraystretch}{1.08}
\begin{tabular}{p{0.23\textwidth}p{0.56\textwidth}r}
\toprule
\textbf{Benchmark} & \textbf{Evaluation scope} & \textbf{Samples} \\
\midrule
MathVerse~\citep{zhang2024mathverse} & Visual grounding under five variants with increasing dependence on diagram information. & 3,940 \\
VisOnlyQA-Syn~\citep{kamoi2025visonlyqa} & Synthetic questions requiring direct perception of geometric relations. & 485 \\
VisOnlyQA-Real~\citep{kamoi2025visonlyqa} & Real-image questions requiring direct geometric perception. & 295 \\
MathVista-GPS~\citep{lu2023mathvista} & Geometry problems from the MathVista benchmark. & 216 \\
\midrule
Geometry3K~\citep{lu2021intergps} & Plane-geometry reasoning over diagrams and textual conditions. & 589 \\
PGPS9K~\citep{zhang2023pgps} & Multimodal plane-geometry problem solving. & 1,000 \\
GeoQA~\citep{chen2021geoqa} & Multimodal numerical reasoning over geometric diagrams. & 754 \\
\midrule
GeoLaux-mini~\citep{fu2026geolaux} & Long-step geometry reasoning requiring auxiliary constructions. & 330 \\
MM-Math~\citep{sun2024mmmath} & Process-level multimodal mathematical reasoning. & 996 \\
\bottomrule
\end{tabular}
\caption{Geometry benchmarks used in evaluation.}
\label{tab:benchmark_summary}
\end{table*}

\subsection{Training and Inference Procedure}
\label{app:procedure}

\noindent\textbf{Model and supervised initialization.}
We initialize Code-CoT from Qwen3-VL-8B-Instruct~\cite{bai2025qwen3vl}.
Supervised fine-tuning is implemented with \texttt{ms-swift} on the $18{,}302$ traces.
We freeze the visual encoder and update the multimodal aligner and language model using a batch size of $64$, a learning rate of $1\times10^{-5}$, and a maximum sequence length of $16{,}384$ for three epochs.
The model is trained to predict the complete Code-CoT response, including perception code, the solution plan, typed reasoning events, and the final answer.

\noindent\textbf{CE-GRPO training.}
CE-GRPO is implemented with \texttt{veRL} and initialized from the Code-CoT SFT checkpoint.
We use full-parameter policy updates with a prompt batch size of $64$, $G=4$ rollouts per prompt, a learning rate of $1\times10^{-6}$, and a rollout temperature of $0.6$, without an additional KL penalty.
Ordinary problem prompts and shared-prefix prompts are mixed at a $1{:}1$ ratio.
For a shared-prefix prompt, the image, question, and preceding Code-CoT prefix are supplied as context, while only the newly generated continuation contributes to the policy loss.

\noindent\textbf{Inference procedure.}
CE-GRPO modifies training only and introduces no additional test-time procedure.
At inference, the model receives the original image and question and generates the perception code, plan, typed events, and final answer autoregressively in a single response.
We use greedy decoding.
The original image remains available throughout generation, and no external geometry solver, second-stage reasoning model, or test-time branching is used.
A protocol-constrained response is considered valid only if it terminates with a non-empty \texttt{<answer>} block.

\subsection{Prompt Templates}
\label{app:prompts}

We use separate prompts for diagram transcription, Code-CoT trace synthesis, and policy generation.
Ordinary and shared-prefix RL examples use the same policy instruction and differ only in whether an assistant prefix is provided.
\Cref{tab:prompt_templates} summarizes the templates; model-specific chat wrappers are omitted.


\section{Experimental Details}
\label{app:experimental_details}

\subsection{Benchmarks and Evaluation Protocol}
\label{app:evaluation}

\noindent\textbf{Benchmarks.}
We evaluate on nine geometry benchmarks spanning visual grounding, plane-geometry reasoning, auxiliary construction, and process-level multimodal reasoning.
\Cref{tab:benchmark_summary} summarizes their evaluation scope and test-set sizes.
Question- and image-level overlaps with the training data are removed as described in~\cref{app:data}.

\noindent\textbf{Evaluation protocol.}
We follow the official evaluation protocols of all benchmarks and report accuracy on each fixed test set.
Unless otherwise stated, all results in the main comparison were obtained through our own end-to-end runs, following the corresponding official implementations and evaluation protocols wherever available; no scores were copied directly from prior reports.
The overall score is the unweighted mean across the nine benchmarks.
For methods following the Code-CoT protocol, a response must terminate with exactly one non-empty \texttt{<answer>} block; otherwise, it is counted as incorrect.

\noindent\textbf{Answer assessment.}
To handle heterogeneous answer formats, Gemini-3.1-Pro-Preview~\cite{google2026gemini31propreview} and Gemini-2.5-Pro~\cite{googledeepmind2025gemini25pro} independently assess whether each prediction is equivalent to the reference answer.
For Code-CoT models, the judges evaluate the extracted content of the final \texttt{<answer>} block.
For baselines using other response formats, they evaluate the complete response.
If the two judgments disagree, the assessment is repeated until both judges reach the same decision.

\begin{table*}[t]
\centering
\scriptsize
\renewcommand{\arraystretch}{1.08}
\setlength{\tabcolsep}{3pt}
\begin{tabular}{p{2.4cm}p{1.8cm}p{2.1cm}p{2.1cm}p{7.4cm}}
\toprule[1pt]
\textbf{Method}
& \textbf{Initialization}
& \textbf{Trainable parameters}
& \textbf{Training hardware}
& \textbf{Core setting and delivered checkpoint} \\
\midrule

Code-CoT SFT
& Qwen3-VL-8B-Instruct
& Language model and aligner; ViT frozen
& $8\times$ A100-80GB
& Batch $64$, learning rate $1\times10^{-5}$, maximum sequence length $16{,}384$; checkpoint $849$. \\

\midrule

DPO
& Code-CoT SFT
& LoRA, $r=64$, $\alpha=128$ ($1.95\%$)
& $8\times$ MI300X
& Effective batch $32$, learning rate $1\times10^{-5}$, $1$ epoch; checkpoint $100$. \\

PPO
& Code-CoT SFT
& Full policy ($100\%$) and a learned critic
& $8\times$ MI300X
& Batch $64$, $4$ rollouts, policy learning rate $1\times10^{-6}$, critic learning rate $2\times10^{-6}$, no KL loss; checkpoint $10$. \\

DAPO
& Code-CoT SFT
& Full policy ($100\%$)
& $8\times$ MI300X
& Batch $64$, $4$ rollouts, learning rate $1\times10^{-6}$; checkpoint $40$. \\

Trajectory-level GRPO
& Code-CoT SFT
& Full model ($100\%$)
& $8\times$ MI300X
& Batch $64$, $8$ rollouts, learning rate $1\times10^{-6}$, temperature $0.6$, no KL loss; checkpoint $90$. \\

\midrule

SRPO-style
& Code-CoT SFT
& Full policy ($100\%$)
& $8\times$ MI300X
& Batch $64$, $4$ rollouts, learning rate $1\times10^{-6}$; checkpoint $60$. \\

GPO adaptation
& Code-CoT SFT
& LoRA, $r=64$, $\alpha=128$ ($1.95\%$)
& $8\times$ MI300X
& Effective batch $32$, learning rate $1\times10^{-5}$, $\beta=0.1$, maximum sequence length $16{,}384$; checkpoint $50$. \\

CFPO-G adaptation
& Code-CoT SFT
& Full policy ($100\%$)
& $8\times$ MI300X
& Batch $64$, $4$ rollouts, learning rate $1\times10^{-6}$, and uniform CMVE with coefficient $0.02$; checkpoint $50$. \\

GRPO-MA adaptation
& Code-CoT SFT
& Full policy ($100\%$)
& $8\times$ MI300X
& Batch $64$, $4$ rollouts, learning rate $1\times10^{-6}$, frozen offline thought branches, and no thought-level credit; checkpoint $50$. \\

\midrule

\rowcolor{gray!15}
\textbf{CE-GRPO}
& Code-CoT SFT
& Full policy ($100\%$)
& $64\times$ A100-80GB
& Batch $64$, $4$ rollouts, learning rate $1\times10^{-6}$, temperature $0.6$, and a $1{:}1$ mix of ordinary and counterfactual prompts drawn from $3{,}270$ fixed prefixes; checkpoint $90$. \\

\bottomrule[1pt]
\end{tabular}
\normalsize
\caption{
Training configurations; percentages denote updated policy parameters, and PPO additionally trains a critic.
}
\label{tab:training_configurations}
\end{table*}

\subsection{Baseline Implementations}
\label{app:baselines}

We compare CE-GRPO with training-free, supervised, general post-training, critical-event and counterfactual post-training, external-checkpoint, and two-stage baselines.

\noindent\textbf{Shared implementation setting.}
All in-house post-training baselines initialize from the same Code-CoT SFT checkpoint and retain the Code-CoT response protocol.
We preserve the original optimization objective of each method while adapting it to the same multimodal geometry setting.
DPO and GPO use LoRA updates, while the remaining post-training baselines use full-parameter updates.
Detailed optimization, hardware, checkpoint, and decoding configurations are reported in~\cref{app:configurations}.

\noindent\textbf{Training-free and supervised baselines.}
The native baseline evaluates Qwen3-VL-8B-Instruct~\citep{bai2025qwen3vl} using its original instruction format.
The prompting baseline applies the Code-CoT instruction in~\cref{app:prompts} to the same frozen backbone without parameter updates.
Code-CoT SFT initializes from Qwen3-VL-8B-Instruct and is trained on the supervised traces described in~\cref{app:data}, without subsequent reinforcement learning.

\noindent\textbf{General post-training baselines.}
We compare with DPO~\citep{rafailov2023dpo}, PPO~\citep{schulman2017ppo}, DAPO~\citep{yu2025dapo}, and trajectory-level GRPO~\citep{shao2024deepseekmath}.
Each method is applied to the shared Code-CoT SFT policy under its original optimization formulation.
Trajectory-level GRPO forms comparison groups from the original image--question prompt and applies one group-relative advantage to the complete generated response.

\noindent\textbf{Critical-event and counterfactual baselines.}
We further compare with SRPO-style~\citep{samanta2026creditassignmentresetslanguage}, GPO~\citep{yu2025gpo}, CFPO~\citep{yu2026cfpo}, and GRPO-MA~\citep{wang2026branching}.
For SRPO-style, we implement the offline self-reset variant used in our experiments.
The other methods retain their respective critical-step, counterfactual, or branching-based optimization mechanisms while operating on Code-CoT responses.
This group provides the closest comparison to CE-GRPO because each method introduces supervision below the complete-trajectory level.

\noindent\textbf{External checkpoints.}
We evaluate the released checkpoints of Qwen2.5-VL-7B-Instruct~\citep{bai2025qwen25vl}, VL-Rethinker-7B~\citep{wang2025vlrethinker}, MMR1-Math-v0-7B~\citep{mmr1team2025}, and G-LLaVA-7B~\citep{gao2023gllava}.
These models are evaluated using their officially released prompt templates, answer formats, and decoding configurations, without adaptation to the Code-CoT protocol.

\noindent\textbf{Two-stage perception--reasoning systems.}
We additionally evaluate GDP-4B-RL~\citep{wang2026geoparsing} and GeoTikzBridge-8B~\citep{sun2026geotikzbridge} as two-stage systems paired with Qwen3-VL-8B.
The first-stage model converts the input diagram into its released intermediate representation, which is then supplied to Qwen3-VL-8B for downstream problem solving.
We use the released prompt template for the first stage and the native Qwen3-VL instruction for the second stage.
These systems require an additional model call, whereas CE-GRPO generates perception code, intermediate reasoning, and the final answer within one response.

\subsection{Comparison Configurations}
\label{app:configurations}

\noindent\textbf{Training configurations.}
\Cref{tab:training_configurations} reports the configuration used to produce each trainable row in the main comparison.
All post-training methods initialize from the same Code-CoT SFT checkpoint.

\begin{table*}[t]
\centering
\footnotesize
\setlength{\tabcolsep}{5pt}
\renewcommand{\arraystretch}{1.10}
\begin{tabularx}{\textwidth}{
p{0.36\textwidth}
p{0.10\textwidth}
X}
\toprule
\textbf{Method or result group}
&
\textbf{Code-CoT protocol}
&
\textbf{Prompt and decoding}
\\
\midrule

Qwen3-VL-8B-Instruct
&
No
&
Native prompt with greedy decoding.
\\

Qwen3-VL-8B-Instruct with Code-CoT prompt
&
Yes
&
Code-CoT prompt with greedy decoding.
\\

Code-CoT SFT
&
Yes
&
Temperature $0.6$, top-$p$ $0.95$, top-$k$ $20$, and repetition penalty $1.05$.
\\

DPO, PPO, DAPO, trajectory-level GRPO, SRPO-style, GPO, CFPO, and GRPO-MA
&
Yes
&
Greedy decoding.
\\

CE-GRPO
&
Yes
&
Greedy decoding.
\\

Qwen2.5-VL-7B-Instruct, VL-Rethinker-7B, and MMR1-Math-v0-7B
&
No
&
Released prompt templates and model-specific decoding.
\\

G-LLaVA-7B
&
No
&
Released prompt template.
\\

GDP-4B-RL $\rightarrow$ Qwen3-VL-8B and GeoTikzBridge-8B $\rightarrow$ Qwen3-VL-8B
&
No
&
Released stage-one prompt followed by the native Qwen3-VL prompt in the second stage.
\\
\bottomrule
\end{tabularx}
\caption{
Inference configurations used in the main evaluation.
External models retain their officially released prompts, decoding settings, and answer formats.
}
\label{tab:inference_configurations}
\end{table*}

\noindent\textbf{Inference configurations.}
We run all in-house inference with \texttt{vLLM} on two A100-80GB GPUs.
\Cref{tab:inference_configurations} reports the answer protocol, prompt, and decoding configuration used for each result group.
Models trained under Code-CoT follow its structured answer protocol, while external checkpoints retain their released prompt templates, output formats, and decoding settings.


\section{Additional Results and Analysis}
\label{app:additional_results}

\subsection{Qualitative Examples}
\label{app:qualitative}

\begin{figure*}[!t]
\centering
\includegraphics[width=\textwidth]{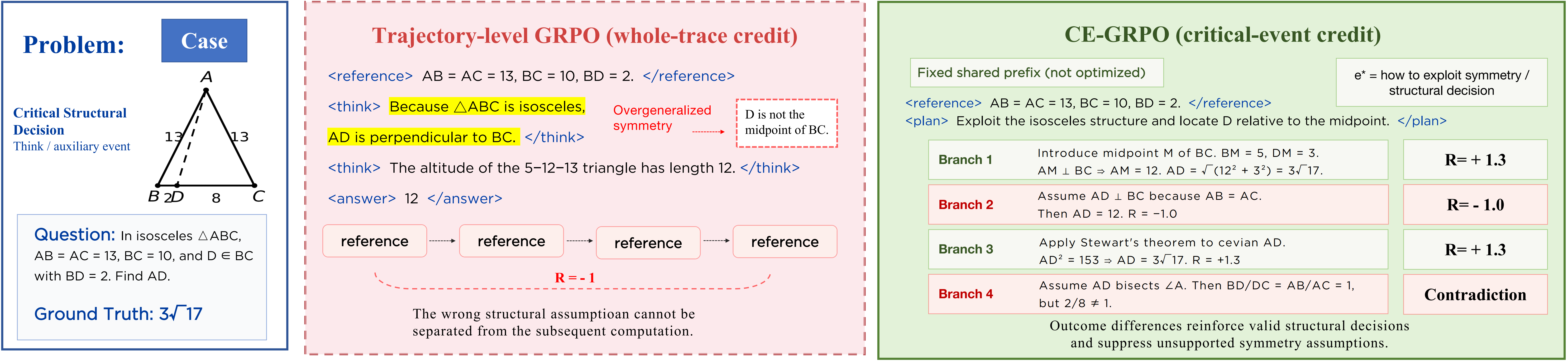}
\caption{
Qualitative comparison between trajectory-level GRPO and CE-GRPO.
Trajectory-level GRPO reinforces an entire incorrect solution that overgeneralizes the symmetry of an isosceles triangle.
CE-GRPO instead branches from the structural decision preceding the error: continuations based on a valid midpoint construction or Stewart's theorem receive positive rewards, whereas unsupported perpendicularity and angle-bisector assumptions are suppressed.
}
\label{fig:case}
\end{figure*}

\noindent\textbf{Trajectory-level credit cannot isolate the structural error.}
\Cref{fig:case} considers an isosceles triangle with $AB=AC=13$, $BC=10$, and $BD=2$, where the correct answer is $AD=3\sqrt{17}$.
The incorrect trajectory overgeneralizes the symmetry of $\triangle ABC$ and assumes that the arbitrary cevian $AD$ is perpendicular to $BC$, yielding $AD=12$.
Trajectory-level GRPO assigns one advantage to the complete response, so the unsupported structural assumption cannot be distinguished from the subsequent deductions that depend on it.

\noindent\textbf{CE-GRPO converts alternative futures into localized credit.}
CE-GRPO instead fixes the prefix preceding this structural decision and samples multiple complete continuations.
Valid branches either introduce the midpoint of $BC$ or apply Stewart's theorem, both recovering $AD=3\sqrt{17}$ and receiving positive rewards.
Branches that assume $AD\perp BC$ or treat $AD$ as an angle bisector produce incorrect or contradictory outcomes.
Because only the regenerated event and its suffix participate in the policy loss, these outcome differences reinforce valid uses of symmetry while suppressing unsupported geometric assumptions.

\subsection{Offline Selector Validation}
\label{app:selector_validation}

\begin{table}[t]
\centering
\small
\begin{tabular}{lccc}
\toprule
Selector & Crit.@2 & $|\Delta R|$ & Tok./critical \\
\midrule
Random
& $0.222$ & $0.327$ & $81.3\mathrm{k}$ \\
Raw entropy
& $0.259$ & $0.376$ & $75.8\mathrm{k}$ \\
Type-normalized entropy
& $0.254$ & $0.374$ & \underline{$75.7\mathrm{k}$} \\
Structure
& $\mathbf{0.289}$ & \underline{$0.405$} & $85.0\mathrm{k}$ \\
Structure + entropy
& \underline{$0.287$} & $\mathbf{0.409}$ & $\mathbf{75.0\mathrm{k}}$ \\
\bottomrule
\end{tabular}
\caption{
Offline selector validation on $300$ problems ($1{,}998$ events per selector).
Crit.@2 is the fraction of traces containing at least one outcome-changing event among two selections; $|\Delta R|$ is the mean absolute reward change; Tok./critical is the token cost per critical event.
Bold and underlined values denote the best and second-best results, respectively; lower Tok./critical is better.
}
\label{tab:selector_validation}
\end{table}

\noindent\textbf{Evaluation protocol.}
We evaluate each selector offline on $300$ problems, with two candidate events selected from each trace.
For each candidate, we preserve the preceding prefix and resample the complete suffix from the same intermediate state.
An event is counted as critical when the original and counterfactual trajectories produce different terminal outcomes.
Accordingly, Crit.@2 measures the fraction of traces containing at least one critical event, $|\Delta R|$ measures the mean absolute reward change, and Tok./critical measures the associated generation cost.

\noindent\textbf{Structure localizes outcome-sensitive events; entropy improves efficiency.}
As shown in~\cref{tab:selector_validation}, entropy-only selectors improve over random selection but remain weaker than structural selection.
The structural prior raises Crit.@2 from $0.222$ to $0.289$, a $30.2\%$ relative improvement, and increases $|\Delta R|$ from $0.327$ to $0.405$, indicating that semantic event boundaries more reliably identify outcome-sensitive branch points.
Adding type-normalized entropy retains a comparable Crit.@2 of $0.287$, achieves the largest reward change of $0.409$, and reduces Tok./critical from $85.0\mathrm{k}$ to $75.0\mathrm{k}$.
These results suggest that structure determines where to branch, while entropy prioritizes informative candidates within the branching budget.
Together with the task-level ablation in~\cref{sec:ablation}, this analysis supports the combined selector adopted by CE-GRPO.

\end{document}